\documentclass{article} 
\usepackage{iclr2016_conference,times}
\usepackage{url}

\title{Variational Auto-encoded Deep Gaussian Processes}

\author{Zhenwen Dai, Andreas Damianou, Javier Gonz\'alez \& Neil Lawrence\\
Department of Computer Science,\\
University of Sheffield, UK\\
\texttt{\{z.dai, andreas.damianou, j.h.gonzalez, n.lawrence\}@sheffield.ac.uk} \\
}

\iclrfinalcopy 

\usepackage{caption}
\usepackage{subcaption}

\usepackage{graphicx} 
\graphicspath{{./},{../diagrams/}}
\DeclareGraphicsExtensions{.pdf,.jpeg,.png}

\usepackage{amsmath}
\usepackage{amssymb}

\global\long\def\dataScalar{y}
\global\long\def\dataVector{\mathbf{\dataScalar}}
\global\long\def\dataMatrix{\mathbf{\MakeUppercase{\dataScalar}}}

\global\long\def\meanVector{\boldsymbol{\mu}}

\global\long\def\inputScalar{x}
\global\long\def\inputMatrix{{\bf \MakeUppercase{\inputScalar}}}
\global\long\def\inputVector{{\bf \inputScalar}}

\global\long\def\covarianceScalar{c}
\global\long\def\covarianceMatrix{\mathbf{\MakeUppercase{\covarianceScalar}}}

\global\long\def\meanScalar{\mu}

\global\long\def\meanVector{\boldsymbol{\meanScalar}}

\global\long\def\mappingFunction{f}

\global\long\def\mappingFunctionMatrix{\mathbf{\MakeUppercase{\mappingFunction}}}

\global\long\def\inducingScalar{u}

\global\long\def\inducingMatrix{\mathbf{\MakeUppercase{\inducingScalar}}}

\global\long\def\inducingInputScalar{z}

\global\long\def\inducingInputMatrix{\mathbf{\MakeUppercase{\inducingInputScalar}}}

\global\long\def\eye{\mathbf{I}}
\global\long\def\identityMatrix{\eye}

\global\long\def\gaussianDist#1#2#3{\mathcal{N}\left(#1|#2,#3\right)}

\global\long\def\expectationDist#1#2{\left\langle #1 \right\rangle _{#2}}

\global\long\def\KL#1#2{\text{KL}\left( #1\,\|\,#2 \right)}

\usepackage{color}
\usepackage{verbatim}

\definecolor{brown}{rgb}{0.9,0.59,0.078}
\definecolor{ironsulf}{rgb}{0,0.7,.5}
\definecolor{lightpurple}{rgb}{0.156,0,0.245}

\ifdefined\blackBackground
\definecolor{colorOne}{rgb}{0, 1, 1}
\definecolor{colorTwo}{rgb}{1, 0, 1}
\definecolor{colorThree}{rgb}{1, 1, 0}
\definecolor{colorTwoThree}{rgb}{1, 0, 0}
\definecolor{colorOneThree}{rgb}{0, 1, 0}
\definecolor{colorOneTwo}{rgb}{0, 0, 1}
\else
\definecolor{colorOne}{rgb}{1, 0, 0}
\definecolor{colorTwo}{rgb}{0, 1, 0}
\definecolor{colorThree}{rgb}{0, 0, 1}
\definecolor{colorTwoThree}{rgb}{0, 1, 1}
\definecolor{colorOneThree}{rgb}{1, 0, 1}
\definecolor{colorOneTwo}{rgb}{1, 1, 0}
\fi

\ifdefined\blackBackground

\global\long\def\blackColor{white}
\global\long\def\whiteColor{black}
\else

\global\long\def\blackColor{black}
\global\long\def\whiteColor{white}
\fi

\global\long\def\bfLambda{\boldsymbol{\Lambda}}

\global\long\def\bfPsi{\boldsymbol{\Psi}}

\global\long\def\bfPhi{\boldsymbol{\Phi}}

\global\long\def\bfg{\mathbf{g}}
\global\long\def\bfh{\mathbf{h}}

\global\long\def\bfx{\mathbf{x}}

\newcommand{\dif}[1]{\text{d}#1}

\global\long\def\eye{\mathbf{I}}

\global\long\def\bfU{\mathbf{U}}
\global\long\def\bfV{\mathbf{V}}
\global\long\def\bfW{\mathbf{W}}

\global\long\def\T{{\top}}
\global\long\def\Tr{\mbox{Tr}}

\global\long\def\cut#1{}

\global\long\def\detail#1{}

\global\long\def\covarianceScalar{k}

\global\long\def\gaussianDist#1#2#3{\mathcal{N}(#1|#2,#3)}

\newcommand{\zM}{\inducingInputMatrix}
\newcommand{\K}{\covarianceMatrix}

\newcommand{\fM}{\mappingFunctionMatrix}

\newcommand{\uM}{\inducingMatrix}

\newcommand{\xM}{\inputMatrix}
\newcommand{\xV}{\inputVector}

\newcommand{\I}{\identityMatrix}
\newcommand{\yM}{\dataMatrix}
\newcommand{\yV}{\dataVector}
\newcommand{\mean}{\meanVector}

\newcommand{\bound}{\mathcal{L}}

\newcommand{\qxMean}{\mean}
\newcommand{\qxCov}{\mathbf{\MakeUppercase{\Sigma}}}

\newcommand{\n}{^{(n)}}
\renewcommand{\l}{_l}

\usepackage{tikz}
\usetikzlibrary{bayesnet}
\usetikzlibrary{positioning,patterns}
\makeatletter
\tikzset{nomorepostaction/.code=\let\tikz@postactions\pgfutil@empty}
\makeatother

\usetikzlibrary{fit}
\tikzset{%
  highlighth/.style={rectangle,rounded corners,fill=red!15,draw,
    fill opacity=0.3,thick,inner sep=0pt}
}
\tikzset{%
  highlightv/.style={rectangle,rounded corners,fill=green!15,draw,
    fill opacity=0.3,thick,inner sep=0pt}
}
\tikzstyle{obs} = [circle,inner sep=1pt,minimum size=20pt, font=\fontsize{10}{10}\selectfont, node distance=1,draw=\blackColor,fill=\blackColor!30]
\tikzstyle{latent} = [obs,fill=\whiteColor]
\tikzstyle{pixel} = [latent,minimum size=10pt, inner sep=0pt,node distance=0,font=\fontsize{5}{5}\selectfont]
\tikzstyle{constObserved} = [latent, node distance=1., fill=\blackColor!30, minimum size=5]
\tikzstyle{constUnobserved} = [latent, node distance=1., fill=\blackColor, minimum size=5]
\tikzstyle{dash plate} = [draw, rectangle, rounded corners, fit=#1]
\tikzstyle{dash plate caption} = [caption, node distance=0, inner sep=0pt,
above=5pt and 0pt of #1.north] %

\begin{document}

\maketitle

\begin{abstract}

We develop a scalable deep non-parametric generative model by augmenting deep Gaussian processes with a recognition model. Inference is performed in a novel scalable variational framework where the variational posterior distributions are reparametrized through a multilayer perceptron. The key aspect of this reformulation is that it prevents the proliferation of variational parameters which otherwise grow linearly in proportion to the sample size. We derive a new formulation of the variational lower bound that allows us to distribute most of the computation in a way that enables to handle datasets of the size of mainstream deep learning tasks. We show the efficacy of the method on a variety of challenges including deep unsupervised learning and deep Bayesian optimization.

\end{abstract}

\section{Introduction}


Probabilistic directed generative models are flexible tools that have recently captured the attention of the the Deep Learning community \citep{UriaEtAl2014,MnihGregor2014,KingmaWelling2013,RezendeEtAl2014}. These models have the ability to produce samples able to mimic the learned data and  they allow principled assessment of the uncertainty in the predictions. These properties are crucial to successfully addressing challenges such as uncertainty quantification or data imputation and allow the ideas of deep learning to be extended to related machine learning fields such as probabilistic numerics.

The main challenge is that exact inference on directed nonlinear probabilistic models is typically intractable due to the required marginalisation of the  latent components. This has lead to the development of probabilistic generative models based on neural networks \citep{KingmaWelling2013, MnihGregor2014, RezendeEtAl2014}, in which probabilistic distributions are defined for the input and output of individual layers. Efficient approximated inference methods have been developed in this context based on stochastic variational inference or stochastic back-propagation. However, a question that remains open is how to properly regularize the model parameters. Techniques such as \emph{dropout} have been used to avoid over-fitting \citep{HintonEtAl2012}. Alternatively, Bayesian inference offers a mathematically grounded framework for regularization. \cite{BlundellEtAl2015} show that Bayesian (variational) inference outperforms dropout. \cite{KingmaEtAl2015, GalGhahramani2015} have shown that dropout itself can be  reformulated in the variational inference context.

In this work, we develop a new scalable Bayesian non-parametric generative model. We focus on a deep Gaussian processes (DGP) that we augment by means of a recognition model, a multi-layer perceptron (MLP) between the latent representation of layers of the DGP. This allows us to simplify the inference and to avoid the challenge of initializing variational parameters. In addition, although DGP have been used only in small scale data so far, we show how it is possible to scale these models by means of new formulation of the lower bound that allows to distribute most of the computation.

The main contributions of this work are: i) a novel extension to DGPs by means of a recognition model that we call \emph{Variational Auto-Encoded deep Gaussian process} (VAE-DGP), ii) a derivation of the distributed variational lower bound of the model and iii) a demonstration of the utility of the model on several mainstream deep learning datasets.

\section{Deep Gaussian Processes}

Gaussian processes provide flexible, non-parametric, probabilistic approaches to function estimation. However, their tractability comes at a price: they can only represent a restricted class of functions. Indeed, even though sophisticated definitions and combinations of covariance functions can lead to powerful models \citep{Durrande:additive11, Gonen:multiple,Hensman:hierarchical13,Duvenaud:structure13,Wilson:gpatt13}, 
the assumption about joint normal distribution of instantiations of the latent function remains; this limits the applicability of the models. 
One line of recent research to address this limitation focused on function composition \citep{Snelson:warped04,Calandra:manifold}. Inspired by deep neural networks, a deep Gaussian process instead employs \emph{process composition} \citep{Lawrence:hgplvm07,Damianou:vgpds11,Lazaro:warped12,Damianou:deepgp13,Hensman:nested14}. 

\begin{figure}[t]
    \begin{center}
    \tikz{%
-        \node[latent] (x) {$\xM_3$};
-        \node[latent, right=2.3 of x] (h1) {$\xM_2$};
        \path[->] 
        (x)    edge  node[sloped, anchor=center, above] {$f_1 \sim \mathcal{GP}$} (h1);
        %
        \node[latent, right=2.3 of h1] (h2) {$\xM_1$};
        \path[->] (h1)    edge  node[sloped, anchor=center, above] {$f_2 \sim \mathcal{GP}$} (h2);
        %
        \node[obs, right=2.3 of h2] (y) {$\yM$};
        \path[->] (h2)    edge  node[sloped, anchor=center, above] {$f_3 \sim \mathcal{GP}$} (y);
        %
        %
    }
    \end{center}
    \caption{A deep Gaussian process with two hidden layers. \label{fig:deepGP1}}
\end{figure}

\begin{figure}[t]
    \centering
    \includegraphics[width=0.9\linewidth]{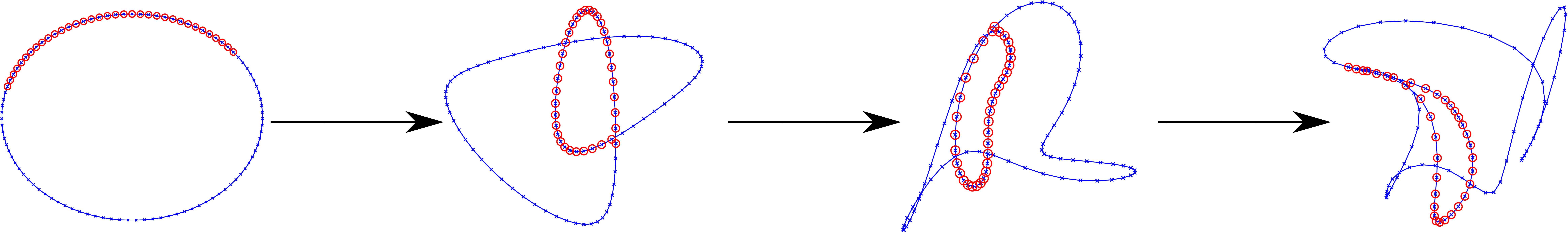}     
    \caption[Samples from a deep GP showing the generation of features.]{Samples from a deep GP showing the generation of features. The upper plot shows the successive non-linear warping of a two-dimensional input space. The red circles correspond to specific locations in the input space for which a feature (a ``loop'') is created in layer 1. As can be seen, as we traverse the hierarchy towards the right, this feature is maintained in the next layers and is potentially further transformed.\label{fig:deepGPSampleWarp}}
\end{figure}

\begin{figure}[t]
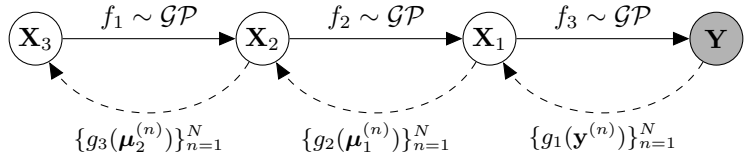

    \begin{center}
    \tikz[]{%
        \node[ latent ] (x) {$\xM_3$};
        \node[ latent, right=2.3 of x ] (h1) {$\xM_2$};
        \path[ -> ] (x) edge  node[sloped, anchor=center, above] {$f_1 \sim \mathcal{GP}$} (h1);
        %
        \node[latent, right=2.3 of h1] (h2) {$\xM_1$};
        \path[->] (h1)    edge  node[sloped, anchor=center, above] {$f_2 \sim \mathcal{GP}$} (h2);
        %
        \node[obs, right=2.3 of h2] (y) {$\yM$};
        \path[->] (h2)    edge  node[sloped, anchor=center, above] {$f_3 \sim \mathcal{GP}$} (y);
        %
        %
        %
        \draw [->, dashed] (y) to [ anchor=center, bend left=60, below] node[label=below:$ \{ g_1( \yV\n)\}_{n=1}^N$] () {} (h2);
       \draw [->, dashed] (h2) to [ anchor=center, bend left=60, below] node[label=below:$\{g_2(  \qxMean\n_{1})\}_{n=1}^N$] () {} (h1);
       \draw [->, dashed] (h1) to [ anchor=center, bend left=60, below] node[label=below:$\{g_3( \qxMean\n_{2})\}_{n=1}^N$] () {} (x);
    }
    \end{center}
    \caption{A deep Gaussian process with three hidden layers and back-constraints. \label{fig:deepGP_backconstrain}}
\end{figure}

A deep GP is a deep directed graphical model that consists of multiple layers of latent variables and employs Gaussian processes to govern the mapping between consecutive layers \citep{Lawrence:hgplvm07,damianou:thesis15}. Observed outputs are placed in the down-most layer and observed inputs (if any) are placed in the upper-most layer, as illustrated in Figure 
\ref{fig:deepGP1}. More formally, consider a set of data $\yM \in \mathbb{R}^{N \times D}$ with $N$ datapoints and $D$ dimensions. A deep GP then defines $L$ layers of latent variables, $\{\xM_l\}_{l=1}^L, \xM_l \in \mathbb{R}^{N \times Q_l}$ through the following nested noise model definition:
\begin{align}
\yM &= f_1(\xM_1) + \epsilon_1, \quad \epsilon_1 \sim \mathcal{N}(0, \sigma^2_1\I)\\
\xM_{l-1} &= f_l(\xM_l) + \epsilon_l, \quad \epsilon_l \sim \mathcal{N}(0, \sigma^2_l\I), \quad l=2\ldots L \label{eq:deepGP}
\end{align}
where the functions $f_l$ are drawn from Gaussian processes with covariance functions $k_l$, i.e.\ $f_l(x) \sim \mathcal{GP}(0,k_l(x,x'))$. In the unsupervised case, the top hidden layer is assigned a unit Gaussian as a fairly uninformative prior which also provides soft regularization, i.e.\ $\xM_L \sim \mathcal{N}(0,\I)$. In the supervised learning scenario, the inputs of the top hidden layer is observed and govern its hidden outputs.



The expressive power of a deep GP is significantly greater than that of a standard GP, because the successive warping of latent variables through the hierarchy allows for modeling non-stationarities and sophisticated, non-parametric functional ``features'' (see Figure \ref{fig:deepGPSampleWarp}). 
Similarly to how a GP is the limit of an infinitely wide neural network, a deep GP is the limit where the parametric function composition of a deep neural network turns into a process composition. Specifically, a deep neural network can be written as:
\begin{equation}
\label{eq:deepGPs_LlayerNN}
\bfg (\bfx)  = \bfV_L^\T \boldsymbol \phi_L(\bfW_{L-1} \boldsymbol \phi_{L-1}(\dots \bfW_2 \boldsymbol \phi_1(\bfU_1\bfx))),
\end{equation}
where $\bfW,\bfU$ and $\bfV$ are parameter matrices and $\boldsymbol \phi(\cdot)$ denotes an activation function. By non-parametrically treating the stacked function composition $g(\bfh)=\bfV^\T \boldsymbol \phi(\bfU \bfh)$ as process composition we obtain the deep GP definition of Equation \ref{eq:deepGP}.


\subsection{Variational Inference}

In a standard GP model, inference is performed by analytically integrating out the latent function $f$. In the DGP case, the latent variables have to additionally be integrated out, to obtain the marginal likelihood of DGPs over the observed data: 
\begin{equation}
p(\yM) = \int p(\yM | \xM_1) \prod_{l=2}^L p(\xM_{l-1} | \xM_l) p(\xM_L) \dif{\xM_1} \ldots \dif{\xM_L}. \label{eqn:marginal_likelihood}
\end{equation} 
 The above marginal likelihood and the following derivation aims at unsupervised learning problems, however, it is straight-forward to extend the formulation to supervised scenario by assuming observed $\xM_L$.
 Bayesian inference in DGPs involves optimizing the model hyper-parameters with respect to the marginal likelihood and inferring the posterior distributions of latent variables for training/testing data.
The exact inference of DGPs is intractable due to the intractable integral in (\ref{eqn:marginal_likelihood}). Approximated inference techniques such as variational inference and EP have been developed \citep{Damianou:deepgp13, BuiEtAl2015}. 
By taking a variational approach, i.e. assuming a variational posterior distribution of latent variables, $q(\{\xM_l\}_{l=1}^L) = \prod_{l=1}^L q(\xM_l)$, a lower bound of the log marginal distribution can be derived as
\begin{align}
\bound = \sum_{l=1}^L \mathcal{F}_l + \sum_{l=1}^{L-1} H(q(\xM_l)) -\KL{q(\xM_L)}{p(\xM_L)}, \label{eqn:bound}
\end{align} 
where $\mathcal{F}_1 = \expectationDist{\log p(\yM | \xM_1)}{q(\xM_1)}$ and $\mathcal{F}_l =\expectationDist{\log p(\xM_{l-1} | \xM_l)}{q(\xM_{l-1})q(\xM_l)}, l=2\ldots L$, are known as \emph{free energy} for individual layers. $H(q(\xM_l))$ denotes the entropy of the variational distribution $q(\xM_l)$ and $\KL{q(\xM_L)}{p(\xM_L)}$ denotes the Kullback-Leibler divergence between $q(\xM_L)$ and $p(\xM_L)$. According to the model definition, both $p(\yM | \xM_1)$ and $p(\xM_{l-1} | \xM_l)$ are Gaussian processes. The variational distribution of $\xM_l$ is typically parameterized as a Gaussian distribution $q(\xM\l) = \prod_{n=1}^N \gaussianDist{\xV\n\l}{\qxMean\n\l}{\qxCov\n\l}$.

\section{Variational Auto-Encoded Model}

\cite{Damianou:deepgp13} provides a tractable variational inference method for DGP by deriving a closed-form lower bound of the marginal likelihood. While successfully demonstrating strengths of DGP, the experiments that they show are limited to very small scales (hundreds of datapoints). The limitation on scalability is mostly due to the computational expensive covariance matrix inversion and the large number of variational parameters (growing linearly with the size of data).


To scale up DGP to handle large datasets, we propose a new deep generative model, by augmenting DGP with a variationally auto-encoded inference mechanism. We refer to this inference mechanism as a \emph{recognition model} (see Figure \ref{fig:deepGP_backconstrain}). A recognition model provides us with a mechanism for constraining the variational posterior distributions of latent variables. Instead of representing variational posteriors as individual variational parameters, which become a big burden to optimization, we define them as a transformation of observed data. This allows us to reduce the number of parameters for optimization (which no longer grow linearly with the size of data) and to perform fast inference at test time. 
A similar constraint mechanism has been referred to as a ``back-constraint'' in the GP literature. \citet{Lawrence:backconstraints06} constrained the latent inputs of a GP with a parametric model to enforce local distance preservation in the inputs; \citet{Ek:ambiguity08} followed the same approach for constraining the latent space with information from additional views of the data. Our formulation differs from the above in that we rather constrain a whole latent posterior distribution through the variational parameters. \citet{Damianou:semidescribed15} also constrained the posterior, but this was achieved using a direct specific parameterization for that distribution, making this back-constraint grow with the number of inputs.
Another difference to the previous approaches is that we consider deep hierarchies of latent spaces and, consequently, of recognition models. Our constraint mechanism is more similar to that of other variationally auto-encoded models, such as \citep{SalakhutdinovHinton2008, SnoekEtAl2012, KingmaWelling2013, MnihGregor2014, RezendeEtAl2014}. The main differences with our work is that are that we have a Bayesian non-parametric generative model and a closed-form variational lower bound. This enables us to be Bayesian when inferring the generative distribution and avoids sampling from variational posterior distributions.

Specifically, for the observed layer, the posterior mean of the variational distribution is defined as a transformation of the observed data:
\begin{equation}
\qxMean\n_1 = g_1( \yV\n),
\end{equation}
where the transformation function $g_1$ is parameterized by a multi-layer perceptron (MLP). Similarly, for the hidden layers, the posterior mean is defined as a transformation of the posterior mean from the lower layer:
\begin{equation}
\qxMean\n\l = g_l( \qxMean\n_{(l-1)}).
\end{equation}
Note that all the transformation functions are deterministic, therefore, the posterior mean of all the hidden layers can be viewed as direct transformations of the observed data, i.e.\ $\qxMean\n\l = g_l (\ldots g_1 (  \yV\n))$. We use the hyperbolic tangent activation function for all the MLPs. The posterior variances $\qxCov\n\l$ are assumed to be diagonal and the same across all the datapoints. 

\begin{figure}
        \centering
    \begin{subfigure}[b]{0.58\textwidth}
        \includegraphics[width=1.\linewidth]{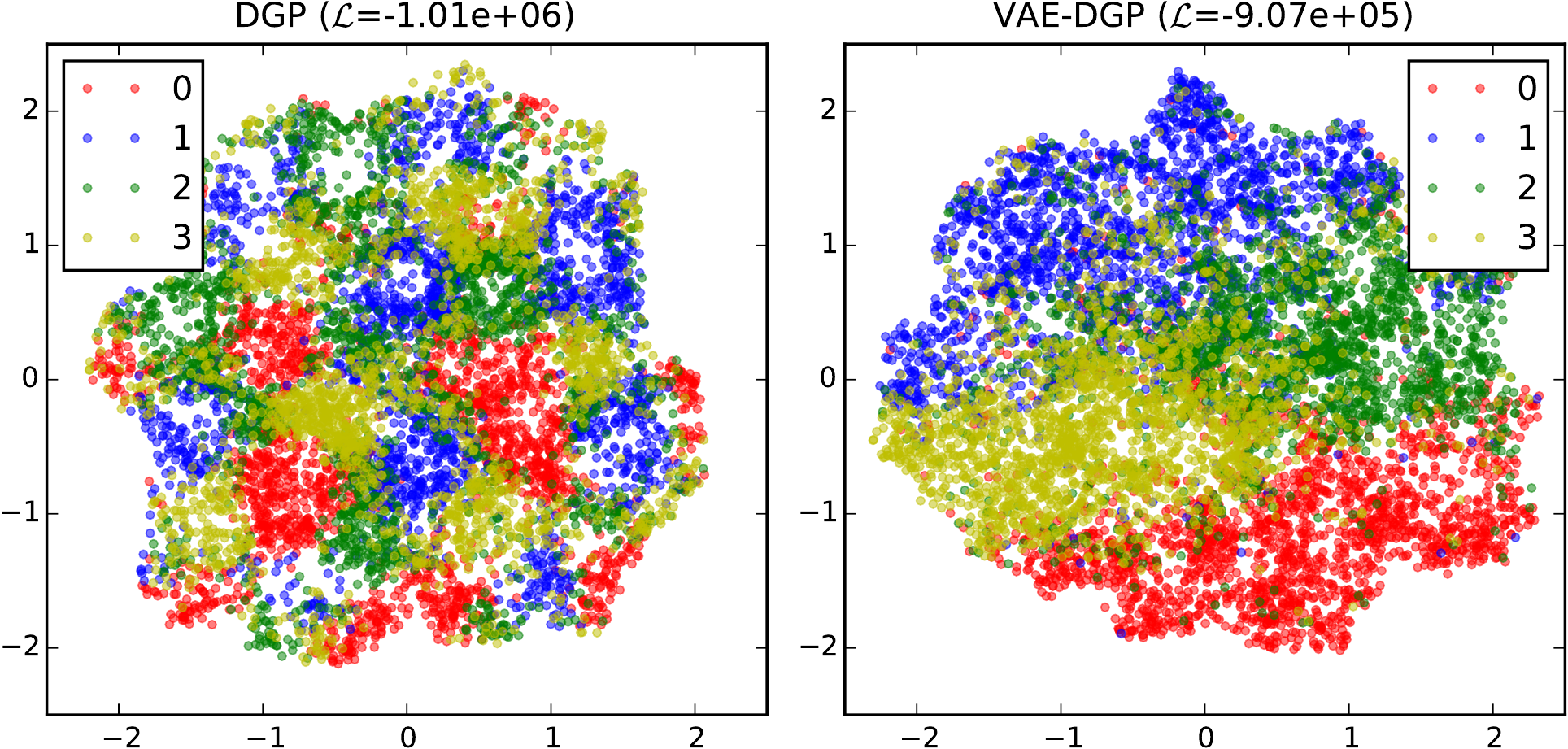}
        \caption{}\label{fig:mnist_noise_comparison}
    \end{subfigure}
    \begin{subfigure}[b]{0.4\textwidth}
    \includegraphics[width=1.\linewidth]{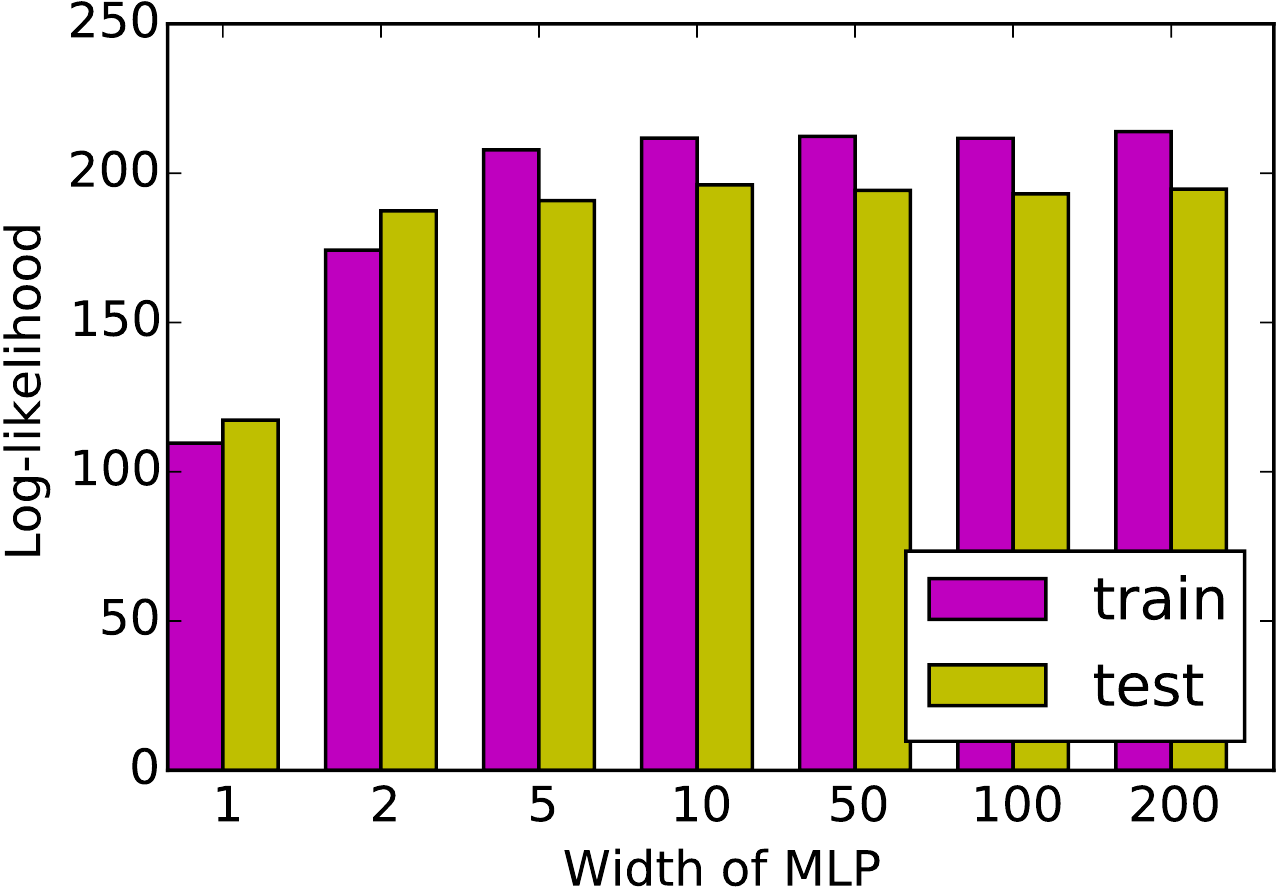}
    \caption{}\label{fig:mnist_VAE_comparison}
    \end{subfigure}
 \caption{(a) The learned 2D latent space of one layer DGP and one layer VAE-DGP from the same initialization on a subset of MNIST with noisy background\protect\footnotemark. A subset of four digits (0, 1, 2, 3) with 10,000 datapoints (2,500 per digit) are taken from the dataset. The left and right figures show the learned 2D latent space on the training data from the same initialization by DGP and VAE-DGP respectively. The recognition model in VAE-DGP helps to avoid local optima and results in a better latent space. (b) The train and test log-likelihood of one layer VAE-DGP with different sizes of recognition models trained on a subset of MNIST (digit ``0" with 1000 for training and 1000 for test). The number of units in the one hidden layer recognition model (MLP) is varied from 1 to 200.}
\end{figure}
\footnotetext{The dataset is downloaded from the link: \url{http://www.iro.umontreal.ca/~lisa/twiki/bin/view.cgi/Public/BackgroundCorrelation}.}

The closed-form variational lower bound allows us to apply sophisticated gradient optimization methods such as L-BFGS. It avoids the problem of initializing and optimizing a large number of variational parameters. The initialization of variational parameters are converted into the initialization of neural network parameters, which has been well studied in deep learning literature. Furthermore, with the reparameterization, the variational parameters are moved coherently during optimization through the changes of neural network mapping. This helps the model avoid local optima and approach better solutions. Figure \ref{fig:mnist_noise_comparison} shows an example of the learned 2D latent space of one layer (shallow) DGP and VAE-DGP from the same initialization. Clearly, the recognition model in VAE-DGP helps move the datapoints to a better solution. Note that the recognition model serves as a (deterministic) reparameterization of variational parameters. Therefore, the parameters of MLP are the variational parameters of our model. As automatically ``regularized" by Bayesian inference, a overly complicated cognition model will not cause the generative model to overfit. This allows us to freely choose a powerful enough recognition model (see Fig.\ \ref{fig:mnist_VAE_comparison} for an example\footnote{Note that the shown training and test log-likelihood are not directly comparable. The shown train log-likelihood is the lower bound in Equation \ref{eqn:bound} divided by the size of data. The shown test log-likelihood is an approximation: $\frac{1}{N_*}(\bound(\yM^*, \yM) -\bound(\yM))$.}).

Computationally, the recognition model re-parameterization resolves the linear growing of the number of variational parameters with respect to the size of data. Based on this formulation, we develop a distributed variational inference approach, which is described in detail in the following section.

\section{Distributed Variational Inference}

The exact evaluation of the variational lower bound in Equation (\ref{eqn:bound}) is still intractable due to the expectation in the free energy terms. A variational approximation technique developed for Bayesian Gaussian Process Latent variable Model (BGPLVM) \citep{Titsias:bayesGPLVM10} can be applied to obtain a lower bound of these free energy terms. Taking the observed layer as an example, by introducing noise-free observations $\fM_1 \in \mathbb{R}^{N \times D}$, a set of auxiliary variable namely inducing variable $\uM_1 \in \mathbb{R}^{M \times D}$ and a set of variational parameter namely inducing inputs $\zM_1 \in \mathbb{R}^{M \times Q_1}$, the conditional distribution is reformulated as
\begin{align}
p(\yM | \xM_1) = \int p(\yM | \fM_1) p(\fM_1 | \uM_1, \xM_1) p(\uM_1) \dif{\fM_1} \dif{\uM_1},
\end{align}
where each row of $\uM_1$ represents an inducing variable which is associated with the inducing input at the same row of $\zM_1$. Assuming a particular form of the variational distribution of $\fM_1$ and $\uM_1$: $q(\fM_1, \uM_1 | \xM_1) = p(\fM_1|\uM_1, \xM_1) q(\uM_1)$, the free energy of the observed layer can be lower bounded by
\begin{align}
\mathcal{F}_1 \geq \expectationDist{\log p(\yM | \fM_1) - \KL{q(\uM_1)}{p(\uM_1)}}{p(\fM_1|\uM_1, \xM_1) q(\uM_1)q(\xM_1)}.  \label{eqn:free-energy1}
\end{align}
As shown by \cite{Titsias:bayesGPLVM10}, this lower bound can be formulated in closed-form for kernels like linear, exponentiated quadratic. For other kernels, it can be computed approximately by using the techniques such as  Gaussian quadrature. Note that the optimal value of $q(\uM_1)$ can be derived in closed-form by setting its gradient $\partial \bound/ \partial q(\uM_1)$ to zero, therefore, the only variational parameters that we need to optimize for the observed layer are $q(\xM_1)$ and $\zM_1$.

For the hidden layers, the variational posterior distributions are slightly different, because the posterior of inducing variables depend on the output variable of that layer. For the $l$-th hidden layer, the variational posterior distribution is, therefore, defined as $q(\fM_l, \uM_l | \xM_{l-1}, \xM_l) = p(\fM_l|\uM_l, \xM_l) q(\uM_l|\xM_{l-1})$.
Similar to the observed layer, a lower bound of the free energy can be derived as:
\begin{align}
\mathcal{F}_l \geq \expectationDist{\log p(\xM_{l-1} | \fM_l) - \KL{q(\uM_l | \xM_{l-1})}{p(\uM_l)}}{p(\fM_l|\uM_l, \xM_l) q(\uM_l | \xM_{l-1})q(\xM_{l-1})q(\xM_l)}.  \label{eqn:free-energyl}
\end{align}
With Equation (\ref{eqn:bound}) and (\ref{eqn:free-energy1}-\ref{eqn:free-energyl}), a closed-form variational lower bound of the log marginal likelihood is defined.

The computation of the lower bounds of free energy terms is expensive. This limits the scalability of the original DGP. Fortunately, with the introduced auxiliary variables and the recognition model, most of the computation is distributable in a data-parallelism fashion. We exploit this fact and derive a distributed formulation of the lower bound. This allows us to scale up our inference method to large data. Specifically, the lower bound of the free energy consists of a few terms (explained below) that depend on the size of data: $\Tr(\yM^\top \yM)$, $\Tr(\bfLambda_1^{-1} \bfPsi_1^\top\yM \yM^\top \bfPsi_1)$, $\psi_1$ and $\bfPhi_1$.
All of them can be formulated as a sum of intermediate results from individual datapoints:
\begin{align*}
\Tr(\yM^\top \yM) &= \sum_{n=1}^N (\yV\n)^\top \yV\n,\\
 \Tr(\bfLambda_1 \bfPsi_1^\top \yM \yM^\top \bfPsi_1) &= \Tr\left(\bfLambda_1^{-1} \left(\sum_{n=1}^N \bfPsi_1\n (\yV\n)^\top\right)\left(\sum_{n=1}^N \bfPsi_1\n (\yV\n)^\top\right)^\top\right),
 \end{align*}
where $\K_{\fM_1\fM_1}$, $\K_{\uM_1\uM_1}$ are the covariance matrices of $\fM_1$ and $\uM_1$ respectively, $\K_{\fM_1\uM_1}$ is the cross-covariance matrix between $\fM_1$ and $\uM_1$, and $\psi_1 = \Tr(\expectationDist{\K_{\fM_1\fM_1}}{q(\xM_1)})$, $\bfPsi_1 = \expectationDist{\K_{\fM_1\uM_1}}{q(\xM_1)}$ and $\bfPhi_1 = \expectationDist{\K_{\fM_1\uM_1}^\top\K_{\fM_1\uM_1}}{q(\xM_1)}$, and $\bfLambda_1 = \K_{\uM_1\uM_1} + \bfPhi_1$. This enables data-parallelism by distributing the computation that depends on individual datapoints and only collecting the intermediate results that do not scale with the size of data. \cite{GalEtAl2014} and \cite{Dai:gpu14} exploit a similar formulation for distributing the computation of BGPLVM, however, in their formulations, the gradients of variational parameters that depend on individual datapoints have to be collected centrally. Such collection severely limits the scalability of the model.

For hidden layers, the free energy terms are slightly different. Their data-dependent terms additionally involve the expectation with respect to the variational distribution of output variables: $\Tr(\expectationDist{\xM_{l-1}^\top \xM_{l-1}}{q(\xM_{l-1})})$, $\Tr(\bfLambda\l^{-1} \bfPsi\l^\top \expectationDist{\xM_{l-1} \xM_{l-1}^\top}{q(\xM_{l-1})} \bfPsi\l)$, $\psi\l$ and $\bfPhi\l$. The first term can be naturally reformulated as a sum across datapoints:
\begin{align}
\Tr(\expectationDist{\xM_{l-1}^\top \xM_{l-1}}{q(\xM_{l-1})}) &= \sum_{n=1}^N (\qxMean\n_{l-1})^\top\qxMean\n_{l-1} +\Tr(\qxCov\n_{l-1}).
\end{align}
For the second term, we can rewrite $ \expectationDist{\xM_{l-1} \xM_{l-1}^\top}{q(\xM_{l-1})} = \mathbf{R}_{l-1}^\top\mathbf{R}_{l-1}+\mathbf{A}_{l-1}\mathbf{A}_{l-1}$, where $\mathbf{R}_{l-1} = [(\qxMean^{(1)}_{(l-1)})^\top,\ldots,(\qxMean^{(N)}_{(l-1)})^\top]$, $\mathbf{A}_{l-1} = \text{diag}(\alpha^{(1)}_{l-1},\ldots,\alpha^{(N)}_{l-1})$ and $\alpha\n_{l-1} = \Tr(\qxCov\n_{l-1})^{\frac{1}{2}}$. This enables us to formulate it into a distributable form:
\begin{align}
\Tr(\bfLambda\l^{-1} \bfPsi\l^\top \expectationDist{\xM_{l-1} \xM_{l-1}}{q(\xM_{l-1})}^\top \bfPsi\l) &= \Tr\left(\bfLambda\l^{-1} \left( \bfPsi\l^\top \mathbf{R}_{l-1}^\top\right)\left( \mathbf{R}_{l-1}\bfPsi\l  \right)\right)\nonumber\\
&+\Tr\left(\bfLambda\l^{-1} \left(\sum_{n=1}^N \bfPsi\l\n \alpha\n_{l-1}\right)\left(\sum_{n=1}^N \bfPsi\l\n \alpha\n_{l-1}\right)^\top\right).
\end{align}
With the above formulations, we obtain distributable a variational lower bound. For optimization, the gradients of all the model and variational parameters can be derived with respect to the lower bound. As the variational distributions $q(\{\xM_l\}_{l=1}^L)$ are computed according to the recognition model, the gradients of $q(\{\xM_l\}_{l=1}^L)$ are back-propagated (through the recognition model), which allows to compute the gradients of its the parameters. 

\section{Experiments}

\begin{figure}[t!]
\centering
    \begin{subfigure}[b]{0.29\textwidth}
        \includegraphics[width=1.\linewidth]{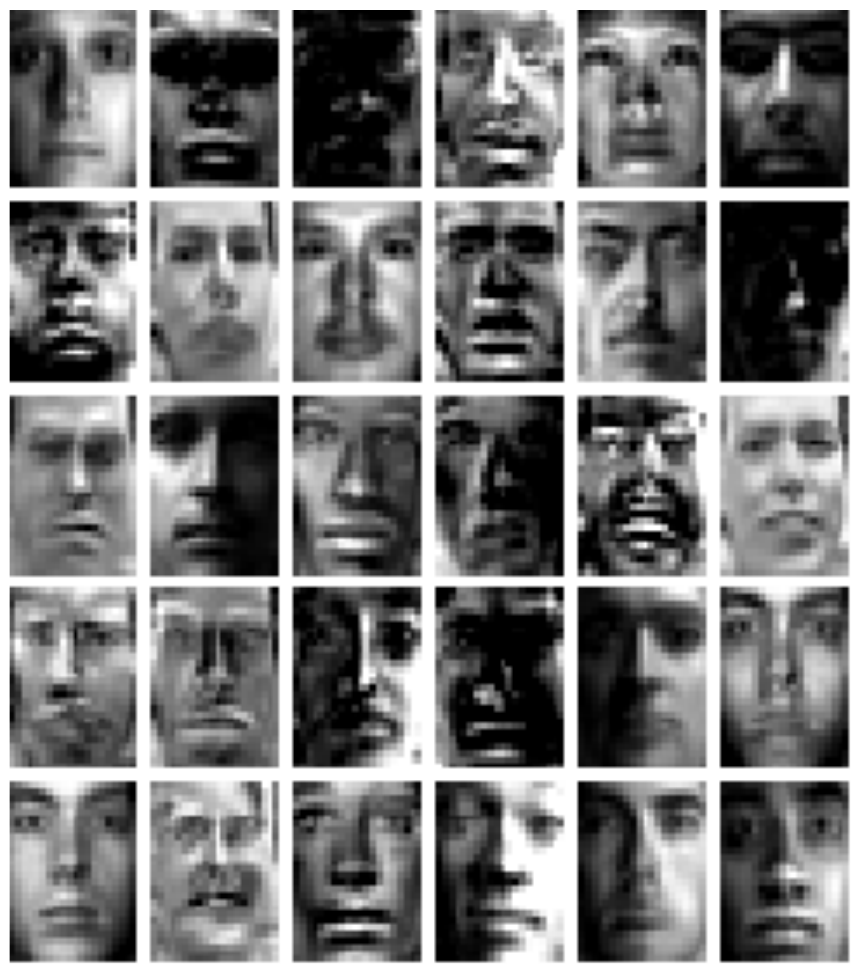}
        \caption{}\label{fig:frey_samples}
    \end{subfigure}
    \begin{subfigure}[b]{0.29\textwidth}
    \includegraphics[width=1.\linewidth]{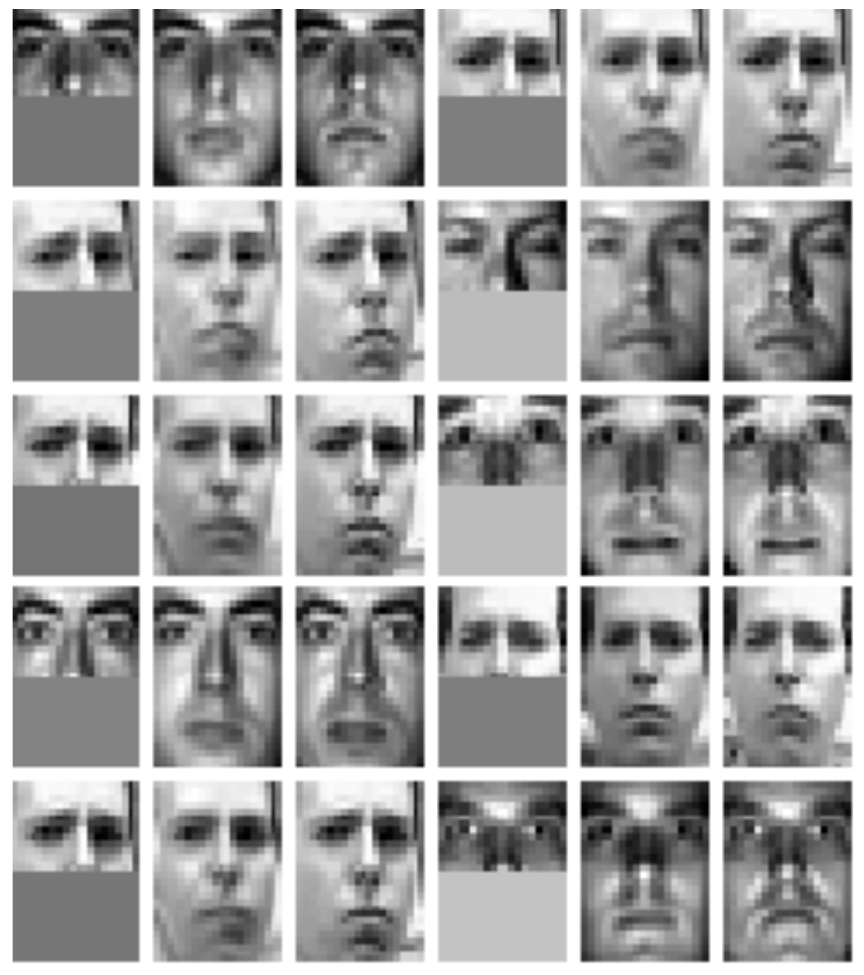}
    \caption{}\label{fig:frey_imputation}
    \end{subfigure}
    \begin{subfigure}[b]{0.395\textwidth}
    \includegraphics[width=1.\linewidth]{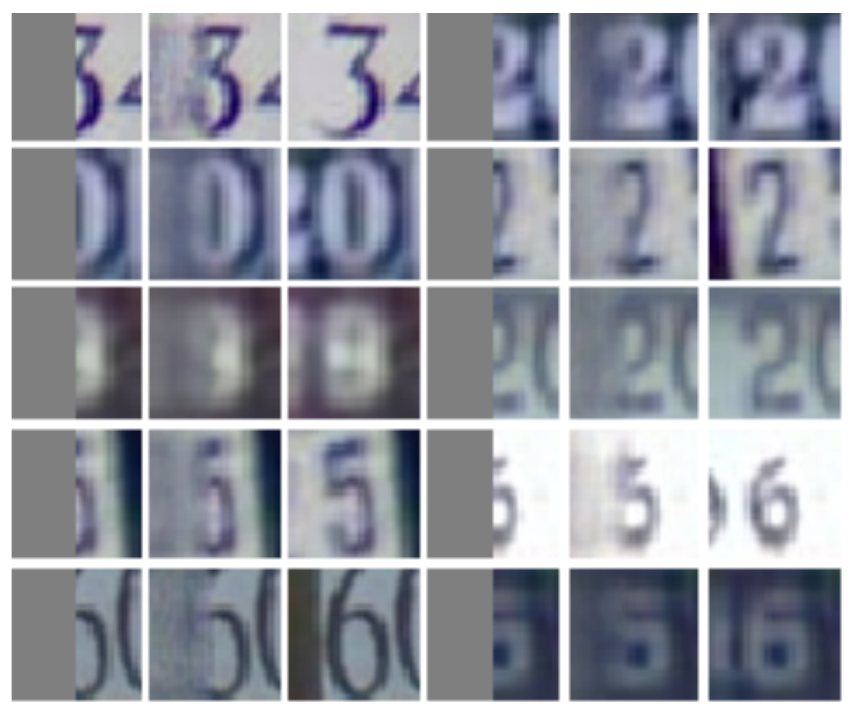}
    \caption{}\label{fig:svhn_imputation}
    \end{subfigure}
 \caption{(a) The samples generated from VAE-DGP trained on the combination of Frey faces and Yale faces (Frey-Yale). (b)  Imputation from the test set of Frey-Yale.  (c) Imputation from the test set of SVHN. The gray color indicates the missing area. The 1st column shows the input images, the 2nd column show the imputed images and 3rd column shows the original full images.}
\end{figure}

As a probabilistic generative model, VAE-DGP is applicable to a range of different tasks such as data generation, data imputation, etc. In this section we evaluate our model in a variety of problems and compare it with the alternatives in the in the literature.



\subsection{Unsupervised Learning}

\begin{minipage}[c]{0.489\textwidth}
\begin{center}
\begin{tabular}{c|c}
\hline
Model & MNIST\\\hline\hline
DBN &138$\pm$2 \\
Stacked CAE & 121 $\pm$ 1.6 \\ 
Deep GSN & 214 $\pm$ 1.1 \\
Adversarial nets & 225 $\pm$ 2 \\
GMMN+AE & 282 $\pm$ 2\\
\hline
VAE-DGP (5) & 301.67 \\
VAE-DGP (10-50) & 674.86 \\
VAE-DGP (5-20-50) & 723.65\\
\hline
\end{tabular}
\captionof{table}{Log-likelihood for the MNIST test data with different models. The baselines are DBN and Stacked CAE \citep{BengioEtAl2013}, Deep GSN \citep{BengioEtAl2014}, Adversarial nets \citep{GoodfellowEtAl2014} and GMMN+AE \citep{LiEtAl2015}.}\label{tab:mnist}
\end{center}
\end{minipage}
\hfill
\begin{minipage}[c]{0.45\textwidth}
        \centering
        \vspace{0pt}
        \includegraphics[width=1\linewidth]{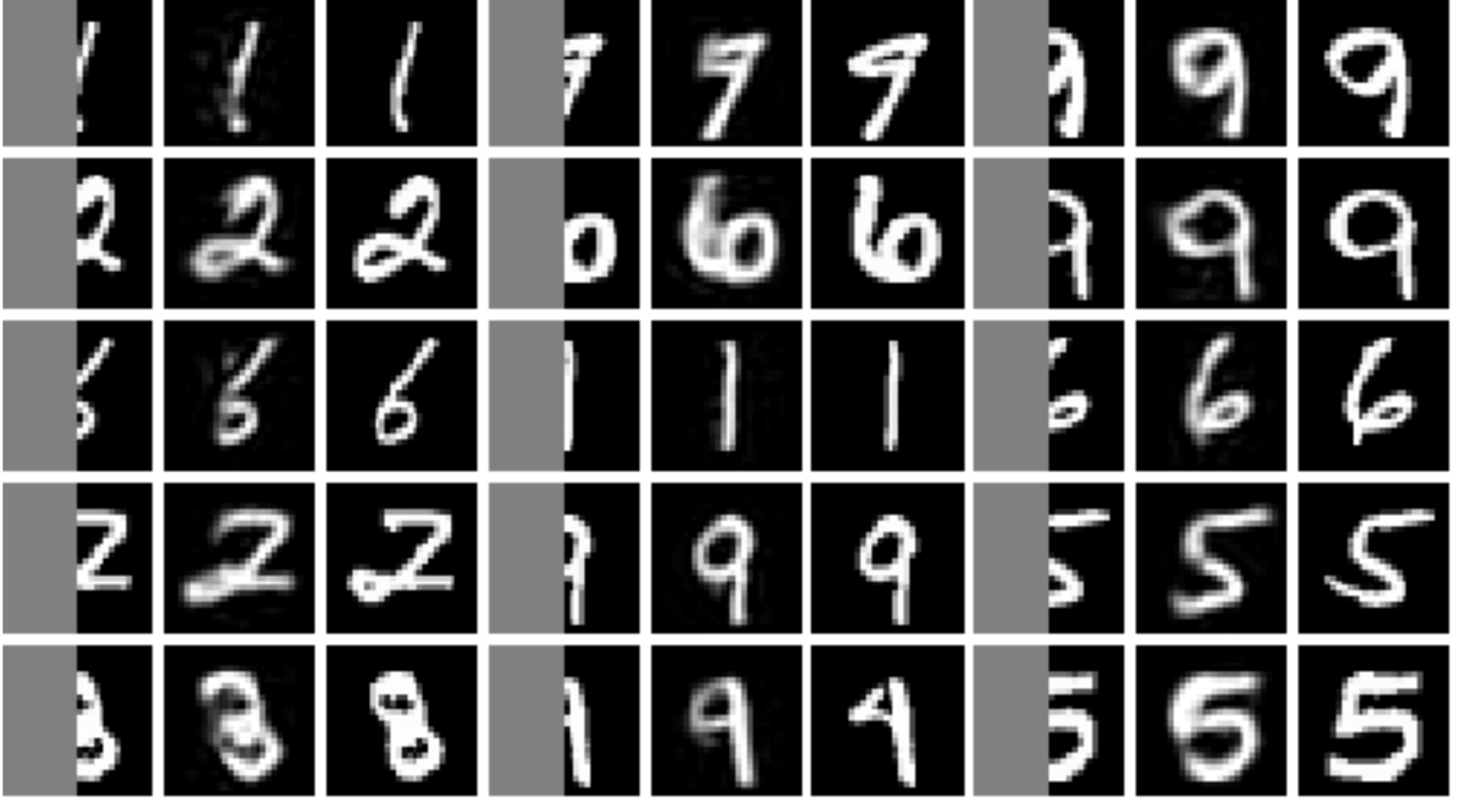}
 \captionof{figure}{Samples of imputation on the test sets. The gray color indicates the missing area. The 1st column shows the input images, the 2nd column show the imputed images and 3rd column shows the original full images.}\label{fig:mnist_imputation}
\end{minipage}

We first apply to our model to the combination of Frey faces and Yale faces (Frey-Yale). The Frey faces contains 1956 $20\times28$ frames taken from a video clip. The Yale faces contains 2414 images, which are resized to $20\times28$. We take the last 200 frames from the Frey faces and 300 images randomly from Yale faces as the test set and use the rest for training. The intensity of the original gray-scale images are normalized to $[0,1]$. The applied VAE-DGP has two hidden layers (a 2D top hidden layer and a 20D middle hidden layer). The exponentiated quadratic kernel is used for all the layers with 100 inducing points. All the MLPs in the recognition model have two hidden layers with widths (500-300). As a generative model, we can draw samples from the learned model by sampling first from the prior distribution of the top hidden layer (a 2D unit Gaussian distribution in this case) and layer-wise downwards. The generated images are shown in Figure \ref{fig:frey_samples}.

To evaluate the ability of our model learning the data distribution, we train the VAE-DGP on MNIST \citep{lecun-98}. We use the whole training set for learning, which consists of 60,000 $28\times28$ images. The intensity of the original gray-scale images are normalized to $[0,1]$. We train our model with three different model settings (one, two and three hidden layers). The trained models are evaluated by the log-likelihood of the test set\footnote{As a non-parametric model, the test log-likelihood of VAE-DGP is formulated as $\frac{1}{N_*}\log p(\yM^*|\yM)$, where $\yM^*$ is the test data and $\yM$ is the training data. As the true test log-likelihood is intractable, we approximate it as $\frac{1}{N_*}(\bound(\yM^*, \yM) -\bound(\yM))$.}, which consists of 10,000 images. The results are shown in Table \ref{tab:mnist} along with some baseline performances taken from the literature. The numbers in the parenthesis indicate the dimensionality of hidden layers from top to bottom. The exponentiated quadratic kernel are used for all the layers with 300 inducing points.  All the MLPs in the recognition model has two hidden layers with width (500-300). All our models are trained as a whole from randomly initialized recognition model.

\subsection{Data Imputation}

We demonstrate the model's ability to impute missing data by showing half of images on the test set. We use the learned VAE-DGP to impute the other half of the images. this is challenging problem because there might be ambiguities in the answers. For instance, by showing the right half of a digit ``8", the answers ``3" and ``8" are both reasonable. We show the imputation performance for the test images in Frey-Yale and MNIST in Fig.\ \ref{fig:frey_imputation} and Fig.\ \ref{fig:mnist_imputation} respectively. We also apply VAE-DGP to the street view house number dataset (SVHN) \citep{svhn}. We use three hidden layers with the dimensionality of latent space from top to bottom (5-30-500). The top two hidden layers use the exponentiated quadratic kernel and the observed layer uses the linear kernel with 500 inducing points. The learned model is used for imputing the images in the test set (see Fig.\ \ref{fig:svhn_imputation}). 

\begin{minipage}[c]{0.489\textwidth}
\begin{center}
     \includegraphics[width=1\linewidth]{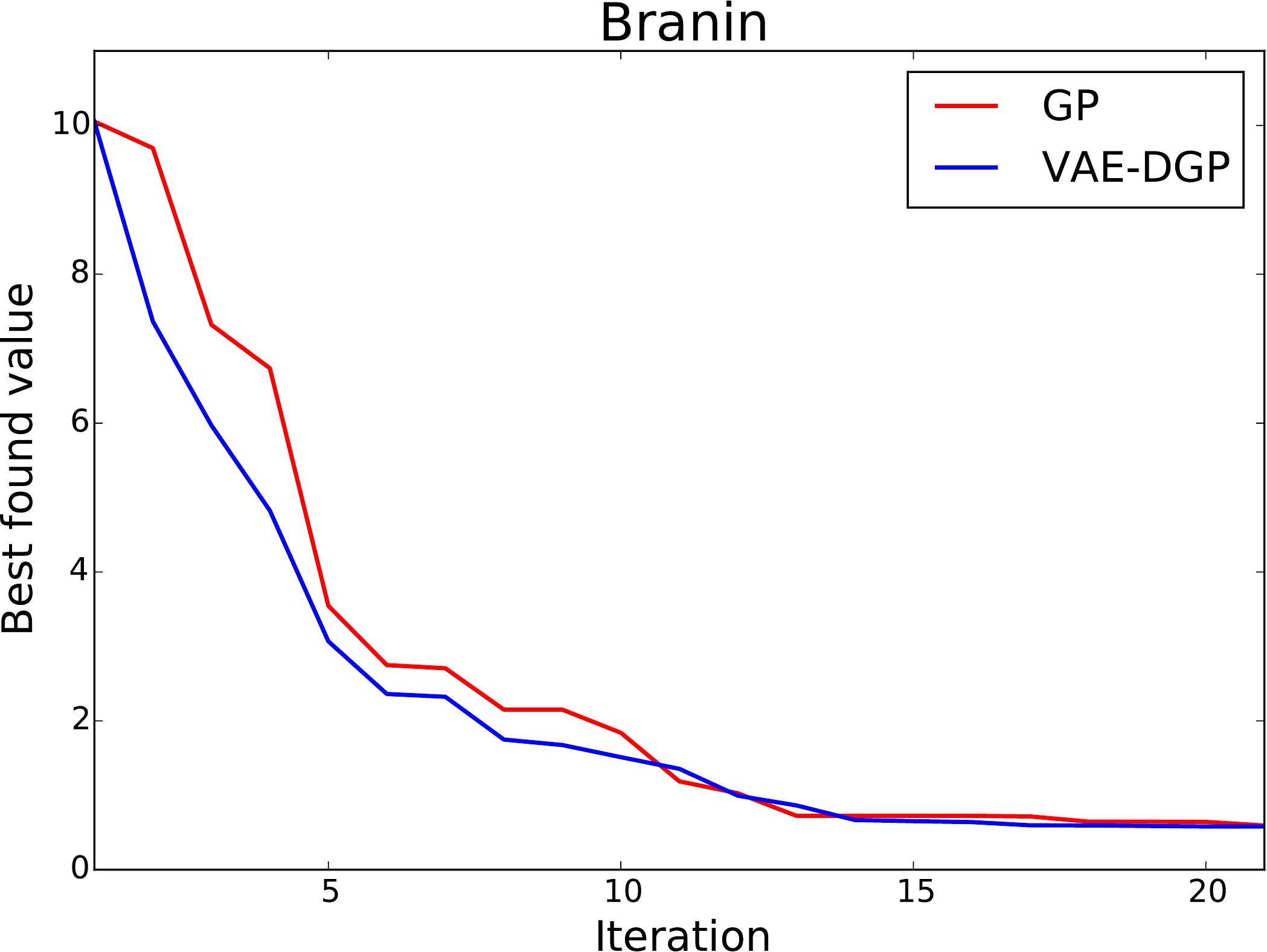} 
\captionof{figure}{Bayesian optimization experiments for the Branin function using a standard GP and our VEA-DGP.}\label{fig:bo}
\end{center}
\end{minipage}
\hfill
\begin{minipage}[c]{0.482\textwidth}
\begin{center}
\begin{tabular}{lc}
\hline
Model & \verb|Abalone| \\
\hline
VEA-DGP     & $\mathbf{825.31} \pm \mathbf{64.35}$ \\
GP          & 888.96 $\pm$ 78.22 \\
Lin. Reg.   & 917.31 $\pm$ 53.76  \\
\hline
\\
\\
\hline 
Model  & \verb|Creep| \\
\hline
VEA-DGP     & $\mathbf{575.39} \pm \mathbf{29.10}$ \\
GP          & 602.11 $\pm$ 29.59 \\
Lin. Reg.   & 1865.76 $\pm$ 23.36 \\
\hline
\\
\end{tabular}
\captionof{table}{MSE obtained from our VEA-DGP, standard GP and linear regression for the Abalone and Creep benchmarks.}\label{table:regression}
\end{center}
\end{minipage}

\subsection{Supervised Learning and Bayesian Optimization}

In this section we consider two supervised learning problem instances: regression and Bayesian optimization (BO) \citep{osborne_bayesian_2010,Snoek*Larochelle*Adams_2012}. We demonstrate the utility of VEA-DGP in these settings by evaluating its performance in terms of predictive accuracy and predictive uncertainty quantification. For these experiments we use a VEA-DGP with one hidden layer (and one observed inputs layer) and exponentiated quadratic covariance functions. Furthermore, we incorporate the deep GP modification of \citet{Duvenaud:pathologies14} so that the observed input layer has an additional connection to the output layer. \citet{Duvenaud:pathologies14} showed that this modification increases the general stability of the method. Since the sample size of the data considered for supervised learning is relatively small, we do not use the recognition model to back-constrain the variational distributions.

In the regression experiments we use the \verb|Abalone| dataset ($4177$ $1$-dimensional outputs and $8-$dimensional inputs) from UCI and the \verb|Creep| dataset ($2066$ $1$-dimensional outputs and $30-$dimensional inputs) from \citep{cole2000modelling}. A typical split for this data is to use $1000$ (Abalone) and $800$ (Creep) instaces for training. We used $100$ inducing inputs for each layer and performed 4 runs with different random splits. We summarize the results in Table \ref{table:regression}.

Next, we show how the VAE-DGP can be used in the context of probabilistic numerics, in particular for Bayesian optimization (BO) \citep{osborne_bayesian_2010,Snoek*Larochelle*Adams_2012}. In BO, the goal is to find  $x_{\text{min}} = \arg \min_{\mathcal{X}} f(x)$ for $\mathcal{X}\subset \mathbb{R}^{Q}$ where a limited number of evaluations are available. Typically, a GP is used to fit the available data, as a surrogate model. The GP is iteratively updated with new function evaluations and used to build an acquisition function able to guide the collection of new observations of $f$. This is done by balancing exploration (regions with large uncertainty) and exploitation (regions with a low mean). In BO, the model is a crucial element of the process: it should be able to express complex classes of functions and to provide coherent estimates of the function uncertainty. In this experiment we use the non-stationary Branin function\footnote{See \url{http://www.sfu.ca/~ssurjano/optimization.html} for details. The default domain is in the experiments.} to compare the performance of standard GPs and the VEA-DPP in the context of BO. We used the popular expected improvement \citep{Jones:1998} acquisition function and we ran 10 replicates of the experiment using different initializations, each kicking-off optimization with 3 points randomly selected from the functions' domain $\mathcal{X}$. In each replicate we iteratively collected 20 evaluations of the functions. In the VEA-DPP we used 30 inducing points. Figure \ref{fig:bo} shows that using the VEA-DPP as a surrogate model results in a gain, especially in the first steps of the optmization. This is due to ability of VEA-DPP to deal with the non-stationary components of the function and to model a much richer class of distributions (e.g.\ multi-modal) in the output layer (as opposed to the standard GP which assumes joint Gaussianity in the outputs).

\section{Conclusion}
We have proposed a new deep non-parametric generative model. Although general enough to be used in supervised and unsupervised problems, we especially highlighted its usefulness in the latter scenario, a case which is known to be a major challenge for current deep machine learning approaches. 
Our model is based on a deep Gaussian process, which we extended with a layer-wise parameterization through multilayer perceptrons, significantly simplifying optimization. Additionally, we developed a new formulation of the lower bound that allows for distributed computations. Overall, our approach is able to perform Bayesian inference using large datasets and compete with current alternatives. Future developments include the regularization of the perceptron weights, to reformulate the current setup for the context of multi-view problems and to incorporate convolutional structures into the objective function.

\textbf{acknowledgement.} The authors thank the financial support of RADIANT (EU FP7-HEALTH Project Ref 305626), BBSRC Project No BB/K011197/1 and WYSIWYD (EU FP7-ICT Project Ref 612139).


{\small
\bibliography{../../../bib/lawrence,../../../bib/other,../../../bib/zbooks,deep_backcstr}

\begin{thebibliography}{42}
\providecommand{\natexlab}[1]{#1}
\providecommand{\url}[1]{\texttt{#1}}
\expandafter\ifx\csname urlstyle\endcsname\relax
  \providecommand{\doi}[1]{doi: #1}\else
  \providecommand{\doi}{doi: \begingroup \urlstyle{rm}\Url}\fi

\bibitem[Bengio et~al.(2013)Bengio, Mesnil, Dauphin, and Rifai]{BengioEtAl2013}
Bengio, Yoshua, Mesnil, Gr{\'{e}}goire, Dauphin, Yann, and Rifai, Salah.
\newblock {Better Mixing via Deep Representations}.
\newblock In \emph{International Conference on Machine Learning}, 2013.

\bibitem[Bengio et~al.(2014)Bengio, Laufer, Alain, and
  Yosinski]{BengioEtAl2014}
Bengio, Yoshua, Laufer, Eric, Alain, Guillaume, and Yosinski, Jason.
\newblock {Deep Generative Stochastic Networks Trainable by Backprop}.
\newblock In \emph{International Conference on Machine Learning}, 2014.

\bibitem[Blundell et~al.(2015)Blundell, Cornebise, Kavukcuoglu, and
  Wierstra]{BlundellEtAl2015}
Blundell, Charles, Cornebise, Julien, Kavukcuoglu, Koray, and Wierstra, Daan.
\newblock {Weight Uncertainty in Neural Networks}.
\newblock In \emph{International Conference on Machine Learning}, 2015.

\bibitem[Bui et~al.(2015)Bui, Hern{\'{a}}ndez-Lobato, Li,
  Hern{\'{a}}ndez-Lobato, and Turner]{BuiEtAl2015}
Bui, Thang~D., Hern{\'{a}}ndez-Lobato, Jos{\'{e}}~Miguel, Li, Yingzhen,
  Hern{\'{a}}ndez-Lobato, Daniel, and Turner, Richard~E.
\newblock {Training Deep Gaussian Processes using Stochastic Expectation
  Propagation and Probabilistic Backpropagation}.
\newblock In \emph{Workshop on Advances in Approximate Bayesian Inference,
  NIPS}, 2015.

\bibitem[Calandra et~al.(2014)Calandra, Peters, Rasmussen, and
  Deisenroth]{Calandra:manifold}
Calandra, Roberto, Peters, Jan, Rasmussen, Carl~Edward, and Deisenroth,
  Marc~Peter.
\newblock Manifold {G}aussian processes for regression.
\newblock Technical report, 2014.

\bibitem[Cole et~al.(2000)Cole, Martin-Moran, Sheard, Bhadeshia, and
  MacKay]{cole2000modelling}
Cole, D, Martin-Moran, C, Sheard, AG, Bhadeshia, HKDH, and MacKay, DJC.
\newblock Modelling creep rupture strength of ferritic steel welds.
\newblock \emph{Science and Technology of Welding \& Joining}, 5\penalty0
  (2):\penalty0 81--89, 2000.

\bibitem[Dai et~al.(2014)Dai, Damianou, Hensman, and Lawrence]{Dai:gpu14}
Dai, Zhenwen, Damianou, Andreas, Hensman, James, and Lawrence, Neil.
\newblock Gaussian process models with parallelization and {GPU} acceleration,
  2014.

\bibitem[Damianou(2015)]{damianou:thesis15}
Damianou, Andreas.
\newblock Deep {G}aussian processes and variational propagation of uncertainty.
\newblock \emph{PhD Thesis, University of Sheffield}, 2015.

\bibitem[Damianou \& Lawrence(2015)Damianou and
  Lawrence]{Damianou:semidescribed15}
Damianou, Andreas and Lawrence, Neil.
\newblock Semi-described and semi-supervised learning with {G}aussian
  processes.
\newblock In \emph{31st Conference on Uncertainty in Artificial Intelligence
  (UAI)}, 2015.

\bibitem[Damianou \& Lawrence(2013)Damianou and Lawrence]{Damianou:deepgp13}
Damianou, Andreas and Lawrence, Neil~D.
\newblock Deep {G}aussian processes.
\newblock In Carvalho, Carlos and Ravikumar, Pradeep (eds.), \emph{Proceedings
  of the Sixteenth International Workshop on Artificial Intelligence and
  Statistics}, volume~31, pp.\  207--215, AZ, USA, 4 2013. JMLR W\&CP 31.

\bibitem[Damianou et~al.(2011)Damianou, Titsias, and
  Lawrence]{Damianou:vgpds11}
Damianou, Andreas, Titsias, Michalis~K., and Lawrence, Neil~D.
\newblock Variational {Gaussian} process dynamical systems.
\newblock In Bartlett, Peter, Peirrera, Fernando, Williams, Chris, and
  Lafferty, John (eds.), \emph{Advances in Neural Information Processing
  Systems}, volume~24, Cambridge, MA, 2011. MIT Press.

\bibitem[Dasgupta \& McAllester(2013)Dasgupta and McAllester]{icml13}
Dasgupta, Sanjoy and McAllester, David (eds.).
\newblock \emph{Proceedings of the 30th International Conference on Machine
  Learning, ICML 2013, Atlanta, GA, USA, 16-21 June 2013}, volume~28 of
  \emph{JMLR Proceedings}, 2013. JMLR.org.

\bibitem[Durrande et~al.(2011)Durrande, Ginsbourger, and
  Roustant]{Durrande:additive11}
Durrande, N, Ginsbourger, D, and Roustant, O.
\newblock Additive kernels for {G}aussian process modeling.
\newblock Technical report, 2011.

\bibitem[Duvenaud et~al.(2013)Duvenaud, Lloyd, Grosse, Tenenbaum, and
  Ghahramani]{Duvenaud:structure13}
Duvenaud, David, Lloyd, James~Robert, Grosse, Roger, Tenenbaum, Joshua~B., and
  Ghahramani, Zoubin.
\newblock Structure discovery in nonparametric regression through compositional
  kernel search.
\newblock In  \citet{icml13}, pp.\  1166--1174.

\bibitem[Duvenaud et~al.(2014)Duvenaud, Rippel, Adams, and
  Ghahramani]{Duvenaud:pathologies14}
Duvenaud, David, Rippel, Oren, Adams, Ryan, and Ghahramani, Zoubin.
\newblock Avoiding pathologies in very deep networks.
\newblock In Kaski, Sami and Corander, Jukka (eds.), \emph{Proceedings of the
  Seventeenth International Workshop on Artificial Intelligence and
  Statistics}, volume~33, Iceland, 2014. JMLR W\&CP 33.

\bibitem[Ek et~al.(2008)Ek, Rihan, Torr, Rogez, and Lawrence]{Ek:ambiguity08}
Ek, Carl~Henrik, Rihan, Jon, Torr, Philip, Rogez, Gregory, and Lawrence,
  Neil~D.
\newblock Ambiguity modeling in latent spaces.
\newblock In {Popescu-Belis}, Andrei and Stiefelhagen, Rainer (eds.),
  \emph{Machine Learning for Multimodal Interaction (MLMI 2008)}, LNCS, pp.\
  62--73. Springer-Verlag, 28--30 June 2008.

\bibitem[Gal \& Ghahramani(2015)Gal and Ghahramani]{GalGhahramani2015}
Gal, Yarin and Ghahramani, Zoubin.
\newblock {Dropout as a Bayesian Approximation: Representing Model Uncertainty
  in Deep Learning}.
\newblock \emph{arXiv:1506.02142}, 2015.

\bibitem[Gal et~al.(2014)Gal, van~der Wilk, and Rasmussen]{GalEtAl2014}
Gal, Yarin, van~der Wilk, Mark, and Rasmussen, Carl~E.
\newblock {Distributed Variational Inference in Sparse Gaussian Process
  Regression and Latent Variable Models}.
\newblock In \emph{Advances in Neural Information Processing System}, 2014.

\bibitem[G\"{o}nen \& Alpaydin(2011)G\"{o}nen and Alpaydin]{Gonen:multiple}
G\"{o}nen, Mehmet and Alpaydin, Ethem.
\newblock Multiple kernel learning algorithms.
\newblock \emph{Journal of Machine Learning Research}, 12:\penalty0 2211--2268,
  Jul 2011.

\bibitem[Goodfellow et~al.(2014)Goodfellow, Pouget-Abadie, Mirza, Xu,
  Warde-Farley, Ozair, Courville, and Bengio]{GoodfellowEtAl2014}
Goodfellow, Ian, Pouget-Abadie, Jean, Mirza, Mehdi, Xu, Bing, Warde-Farley,
  David, Ozair, Sherjil, Courville, Aaron, and Bengio, Yoshua.
\newblock {Generative Adversarial Networks}.
\newblock In \emph{Advances in Neural Information Processing Systems}, 2014.

\bibitem[Hensman \& Lawrence(2014)Hensman and Lawrence]{Hensman:nested14}
Hensman, James and Lawrence, Neil~D.
\newblock Nested variational compression in deep {G}aussian processes.
\newblock Technical report, University of Sheffield, 2014.

\bibitem[Hensman et~al.(2013)Hensman, Lawrence, and
  Rattray]{Hensman:hierarchical13}
Hensman, James, Lawrence, Neil~D., and Rattray, Magnus.
\newblock Hierarchical {B}ayesian modelling of gene expression time series
  across irregularly sampled replicates and clusters.
\newblock \emph{BMC Bioinformatics}, 14\penalty0 (252), 2013.
\newblock \doi{doi:10.1186/1471-2105-14-252}.

\bibitem[Hinton et~al.(2012)Hinton, Srivastava, Krizhevsky, Sutskever, and
  Salakhutdinov]{HintonEtAl2012}
Hinton, Geoffrey~E., Srivastava, Nitish, Krizhevsky, Alex, Sutskever, Ilya, and
  Salakhutdinov, Ruslan~R.
\newblock {Improving neural networks by preventing co-adaptation of feature
  detectors}.
\newblock \emph{arXiv: 1207.0580}, 2012.

\bibitem[Jones et~al.(1998)Jones, Schonlau, and Welch]{Jones:1998}
Jones, Donald~R., Schonlau, Matthias, and Welch, William~J.
\newblock Efficient global optimization of expensive black-box functions.
\newblock \emph{Journal of Global Optimization}, 13\penalty0 (4):\penalty0
  455--492, 1998.

\bibitem[Kingma \& Welling(2013)Kingma and Welling]{KingmaWelling2013}
Kingma, Diederik~P and Welling, Max.
\newblock {Auto-Encoding Variational Bayes}.
\newblock In \emph{ICLR}, 2013.

\bibitem[Kingma et~al.(2015)Kingma, Salimans, and Welling]{KingmaEtAl2015}
Kingma, Diederik~P., Salimans, Tim, and Welling, Max.
\newblock {Variational Dropout and the Local Reparameterization Trick}.
\newblock In \emph{Advances in Neural Information Processing System}, 2015.

\bibitem[Lawrence \& Moore(2007)Lawrence and Moore]{Lawrence:hgplvm07}
Lawrence, Neil~D. and Moore, Andrew~J.
\newblock Hierarchical {G}aussian process latent variable models.
\newblock In Ghahramani, Zoubin (ed.), \emph{Proceedings of the International
  Conference in Machine Learning}, volume~24, pp.\  481--488. Omnipress, 2007.
\newblock ISBN 1-59593-793-3.

\bibitem[Lawrence \& {Qui\~nonero Candela}(2006)Lawrence and {Qui\~nonero
  Candela}]{Lawrence:backconstraints06}
Lawrence, Neil~D. and {Qui\~nonero Candela}, Joaquin.
\newblock Local distance preservation in the {GP-LVM} through back constraints.
\newblock In Cohen, William and Moore, Andrew (eds.), \emph{Proceedings of the
  International Conference in Machine Learning}, volume~23, pp.\  513--520.
  Omnipress, 2006.
\newblock ISBN 1-59593-383-2.
\newblock \doi{10.1145/1143844.1143909}.

\bibitem[L{\'a}zaro-Gredilla(2012)]{Lazaro:warped12}
L{\'a}zaro-Gredilla, Miguel.
\newblock {B}ayesian warped {G}aussian processes.
\newblock In Bartlett, Peter~L., Pereira, Fernando C.~N., Burges, Christopher
  J.~C., Bottou, L{\'e}on, and Weinberger, Kilian~Q. (eds.), \emph{Advances in
  Neural Information Processing Systems}, volume~25, Cambridge, MA, 2012.

\bibitem[LeCun et~al.(1998)LeCun, Bottou, Bengio, and Haffner]{lecun-98}
LeCun, Y., Bottou, L., Bengio, Y., and Haffner, P.
\newblock Gradient-based learning applied to document recognition.
\newblock \emph{Proceedings of the IEEE}, 86\penalty0 (11):\penalty0
  2278--2324, November 1998.

\bibitem[Li et~al.(2015)Li, Swersky, and Zemel]{LiEtAl2015}
Li, Yujia, Swersky, Kevin, and Zemel, Richard.
\newblock {Generative Moment Matching Networks}.
\newblock In \emph{International Conference on Machine Learning}, 2015.

\bibitem[Mnih \& Gregor(2014)Mnih and Gregor]{MnihGregor2014}
Mnih, A. and Gregor, K.
\newblock {Neural Variational Inference and Learning in Belief Networks}.
\newblock In \emph{International Conference on Machine Learning}, 2014.

\bibitem[Netzer et~al.(2011)Netzer, Wang, Coates, Bissacco, Wu, and Ng]{svhn}
Netzer, Yuval, Wang, Tao, Coates, Adam, Bissacco, Alessandro, Wu, Bo, and Ng,
  Andrew~Y.
\newblock {Reading Digits in Natural Images with Unsupervised Feature
  Learning}.
\newblock \emph{NIPS Workshop on Deep Learning and Unsupervised Feature
  Learning}, 2011.

\bibitem[Osborne(2010)]{osborne_bayesian_2010}
Osborne, Michael.
\newblock \emph{Bayesian {Gaussian} Processes for Sequential Prediction,
  Optimisation and Quadrature}.
\newblock PhD thesis, University of Oxford, 2010.

\bibitem[Rezende et~al.(2014)Rezende, Mohamed, and Wierstra]{RezendeEtAl2014}
Rezende, D~J, Mohamed, S, and Wierstra, D.
\newblock {Stochastic backpropagation and approximate inference in deep
  generative models}.
\newblock In \emph{International Conference on Machine Learning}, 2014.

\bibitem[Salakhutdinov \& Hinton(2008)Salakhutdinov and
  Hinton]{SalakhutdinovHinton2008}
Salakhutdinov, Ruslan and Hinton, Geoffrey.
\newblock {Using Deep Belief Nets to Learn Covariance Kernels for Gaussian
  Processes}.
\newblock In \emph{Advances in Neural Information Processing Systems},
  volume~20, 2008.

\bibitem[Snelson et~al.(2004)Snelson, Rasmussen, and
  Ghahramani]{Snelson:warped04}
Snelson, Edward, Rasmussen, Carl~Edward, and Ghahramani, Zoubin.
\newblock Warped {G}aussian processes.
\newblock In Thrun, Sebastian, Saul, Lawrence, and Sch\"olkopf, Bernhard
  (eds.), \emph{Advances in Neural Information Processing Systems}, volume~16,
  Cambridge, MA, 2004. MIT Press.

\bibitem[Snoek et~al.(2012{\natexlab{a}})Snoek, Adams, and
  Larochelle]{SnoekEtAl2012}
Snoek, Jasper, Adams, Ryan~P., and Larochelle, Hugo.
\newblock {Nonparametric Guidance of Autoencoder Representations using Label
  Information}.
\newblock \emph{Journal of Machine Learning Research}, 13:\penalty0 2567--2588,
  2012{\natexlab{a}}.

\bibitem[Snoek et~al.(2012{\natexlab{b}})Snoek, Larochelle, and
  Adams]{Snoek*Larochelle*Adams_2012}
Snoek, Jasper, Larochelle, Hugo, and Adams, Ryan~P.
\newblock \emph{Practical {B}ayesian optimization of machine learning
  algorithms}, pp.\  2951–2959.
\newblock 2012{\natexlab{b}}.

\bibitem[Titsias \& Lawrence(2010)Titsias and Lawrence]{Titsias:bayesGPLVM10}
Titsias, Michalis~K. and Lawrence, Neil~D.
\newblock Bayesian {G}aussian process latent variable model.
\newblock In Teh, Yee~Whye and Titterington, D.~Michael (eds.),
  \emph{Proceedings of the Thirteenth International Workshop on Artificial
  Intelligence and Statistics}, volume~9, pp.\  844--851, Chia Laguna Resort,
  Sardinia, Italy, 13-16 May 2010. JMLR W\&CP 9.

\bibitem[Uria et~al.(2014)Uria, Murray, and Larochelle]{UriaEtAl2014}
Uria, Benigno, Murray, Iain, and Larochelle, Hugo.
\newblock {A Deep and Tractable Density Estimator}.
\newblock In \emph{International Conference on Machine Learning}, 2014.

\bibitem[Wilson \& Adams(2013)Wilson and Adams]{Wilson:gpatt13}
Wilson, Andrew~Gordon and Adams, Ryan~Prescott.
\newblock Gaussian process kernels for pattern discovery and extrapolation.
\newblock In  \citet{icml13}, pp.\  1067--1075.

\end{thebibliography}
\bibliographystyle{iclr2016_conference}
}

\end{document}